\documentclass[runningheads]{llncs}
\usepackage{graphicx}
\usepackage[dvipsnames,table,xcdraw]{xcolor}
\usepackage{tikz}
\usepackage{comment}
\usepackage{amsmath,amssymb} 
\usepackage{color}
\usepackage{bbding}
\usepackage{multirow}
\usepackage{wrapfig}
\usepackage{makecell}
\usepackage{lscape}
\usepackage[breaklinks=true,colorlinks,bookmarks=false]{hyperref}
\usepackage{caption}
\usepackage{subcaption}

\usepackage[accsupp]{axessibility}  

\begin{document}
\pagestyle{headings}
\mainmatter
\def\ECCVSubNumber{100}  

\title{You Only Need a Good Embeddings Extractor to Fix Spurious Correlations
} 

\titlerunning{You only need a good embeddings extractor to fix spurious correlations} 
\authorrunning{Mehta et al.} 
\author{Raghav Mehta \quad  Vítor Albiero \quad Li Chen \quad Ivan Evtimov \quad Tamar Glaser \\
Zhiheng Li \quad Tal Hassner}
\institute{Meta AI}

\maketitle

\begin{abstract}
Spurious correlations in training data often lead to robustness issues since models learn to use them as shortcuts. For example, when predicting whether an object is a cow, a model might learn to rely on its green background, so it would do poorly on a cow on a sandy background
A standard dataset for measuring state-of-the-art on methods mitigating this problem is Waterbirds. The best method (Group Distributionally Robust Optimization - GroupDRO) currently achieves 89\% worst group accuracy and standard training from scratch on raw images only gets 72\%. GroupDRO requires training a model in an end-to-end manner with subgroup labels. In this paper, we show that we can achieve up to 90\% accuracy without using any sub-group information in the training set by simply using embeddings from a large pre-trained vision model extractor and training a linear classifier on top of it. With experiments on a wide range of pre-trained models and pre-training datasets, we show that the capacity of the pre-training model and the size of the pre-training dataset matters. Our experiments reveal that high capacity vision transformers perform better compared to high capacity convolutional neural networks, and larger pre-training dataset leads to better worst-group accuracy on the spurious correlation dataset. 

\end{abstract}

\section{Introduction}
Many machine learning models may rely on spurious correlation prevalent in the training dataset for learning classification boundaries~\cite{alcorn2019strike,geirhos2020shortcut}. When spurious correlations are not present in real world (validation or testing) datasets, model performance is severely degraded~\cite{arjovsky2019invariant,beery2018recognition}.
Most methods considered in the literature so far mitigate this robustness gap by using explicit or inferred labels of spuriously correlated subsets in the training data.
Here, we observe that training simple linear classifiers on embeddings extracted from frozen, large pre-trained networks might be enough to mitigate spurious correlations. 

To study this phenomenon, we worked with Waterbirds dataset~\cite{sagawa2019distributionally}, one of the common benchmarks for tracking state of the art on mitigation of spurious correlations.
The task is to classify whether an image shows a waterbird or a landbird. In this dataset, each bird is labeled as one of waterbird or landbird and is placed on one of water background (BG) or land background. In the training dataset, the majority (95\%) of waterbirds are placed on a water background, while the majority (95\%) of landbirds are placed on a land background. This leads to background being spuriously correlated with actual class labels. Training a classifier with \textit{Empirical Risk Minimization} (ERM)~\cite{vapnik1991principles} loss on the training dataset could lead to a biased classifier that performs poorly on minority groups (waterbirds on land background and landbirds on water background) at test time. Instead of learning the actual classification between waterbirds or landbirds, the classifier relies on spurious correlation of background with actual labels. A good classifier should perform well on minority groups without relying on the spurious correlations. This is tracked by measuring the ``worst-group accuracy'' (WGA) across the majority and the minority groups in the test data. 

Most methods in the literature use explicit or inferred labels of the majority and minority groups in the training data to achieve high WGA. GroupDRO~\cite{sagawa2019distributionally} minimizes the maximum loss across different groups in the training dataset. \textit{Just Train Twice} (JTT)~\cite{liu2021just} trains a model for a small amount of epochs, up-weights the samples where this model made mistakes, and trains another model using these up-weighted samples. \textit{Subsampling large Groups} (SUBG)~\cite{idrissi2022simple} trains a model with data balanced set across different groups by subsampling images from the majority classes to match the frequency of the minority class. \textit{Deep Feature Reweighting} DFR~\cite{kirichenko2022last} trains a model with ERM loss, and following that, replaces the classifier with another classifier which is trained with a group-balanced dataset, similar to SUBG~\cite{idrissi2022simple}. Other methods include \textit{Invariant Risk Minimization} (IRM)~\cite{arjovsky2019invariant}, \textit{Learning from Failure} (LfF)~\cite{nam2020learning}, \textit{Environment Inference for Invariant Learning} (EIIL)~\cite{creager2021environment}, and \textit{Entangling and Disentangling} (EnD) Deep Representations for Bias Correction~\cite{tartaglione2021end}.

Critically, in all of the above cases, subset labels are either given explicitly in the training data 
or inferred during the model design and training 
. However, this approach has several limitations. First, explicit labels are often not available in real-world scenarios due to privacy concerns, and instances could be part of multiple sub-groups, if different cues are present in a single instance. Second, using even inferred labels for the subgroup may not be accurate, and can be problematic if they are sensitive. Therefore, we believe information about the subgroups should be used only for validation and testing and not for training.

Some recent studies \cite{shi2022robust,kim2022broad} show that good pre-training and better architectures can be useful for domain generalization \cite{kim2022broad} and distribution shifts \cite{kim2022broad}.  In this work, we observe that pre-training also helps with mitigating model shortcuts due to spurious correlations in the training data. All of the limitations of using subgroup labels may be overcome by not using them and instead relying on embeddings from large, pre-trained feature-extracting networks. In our experiments on Waterbirds, we achieve state-of-the-art WGA simply by freezing the feature extractor and training simple linear classifiers on the embeddings extracted from it.


\begin{figure}
    \centering
    \includegraphics[width=\textwidth]{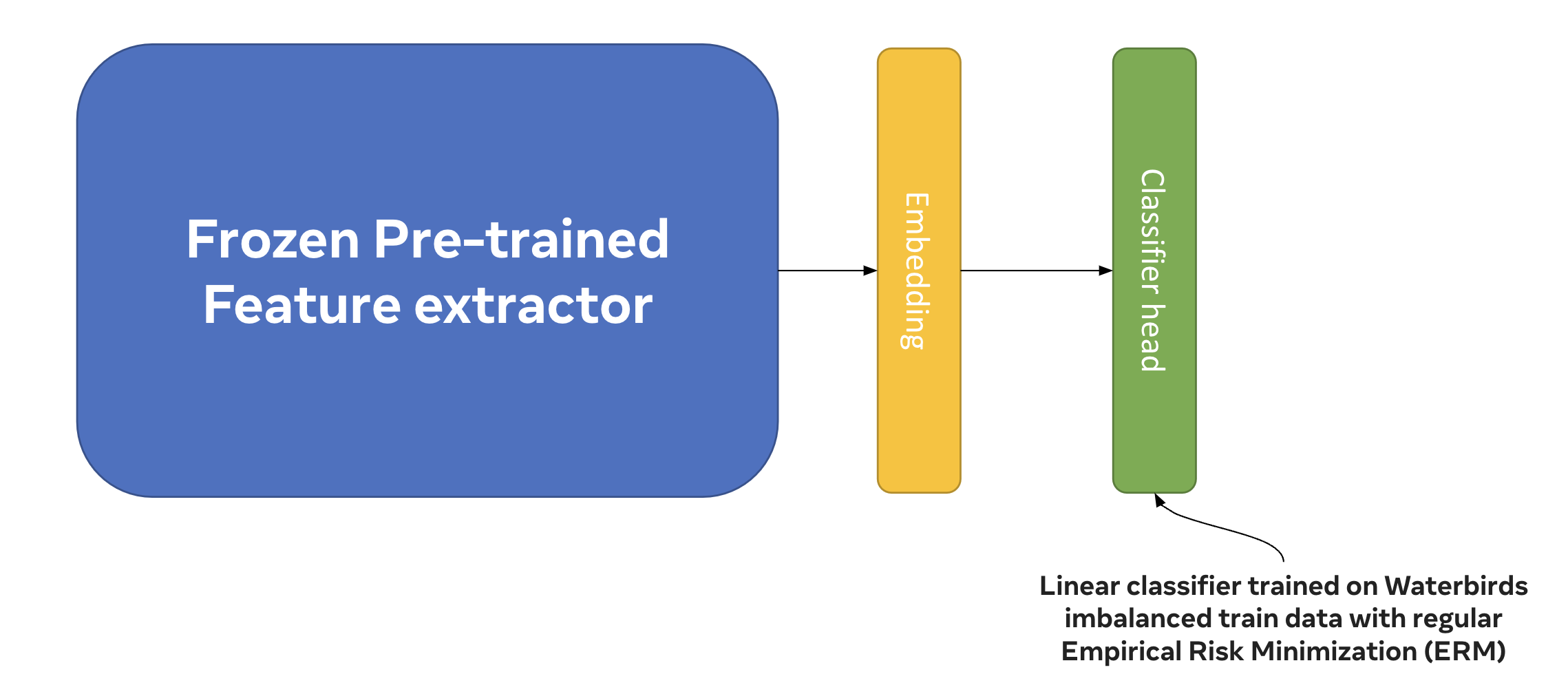}
    \caption{Overview of our proposed method. We use frozen pre-trained models as embedding extractors and only train a linear classifier layer on top of these embeddings with a standard Empirical Risk Minimization (ERM) loss.}
    \label{fig:my_label}
\end{figure}

\begin{figure}
    \centering
    \includegraphics[width=\textwidth]{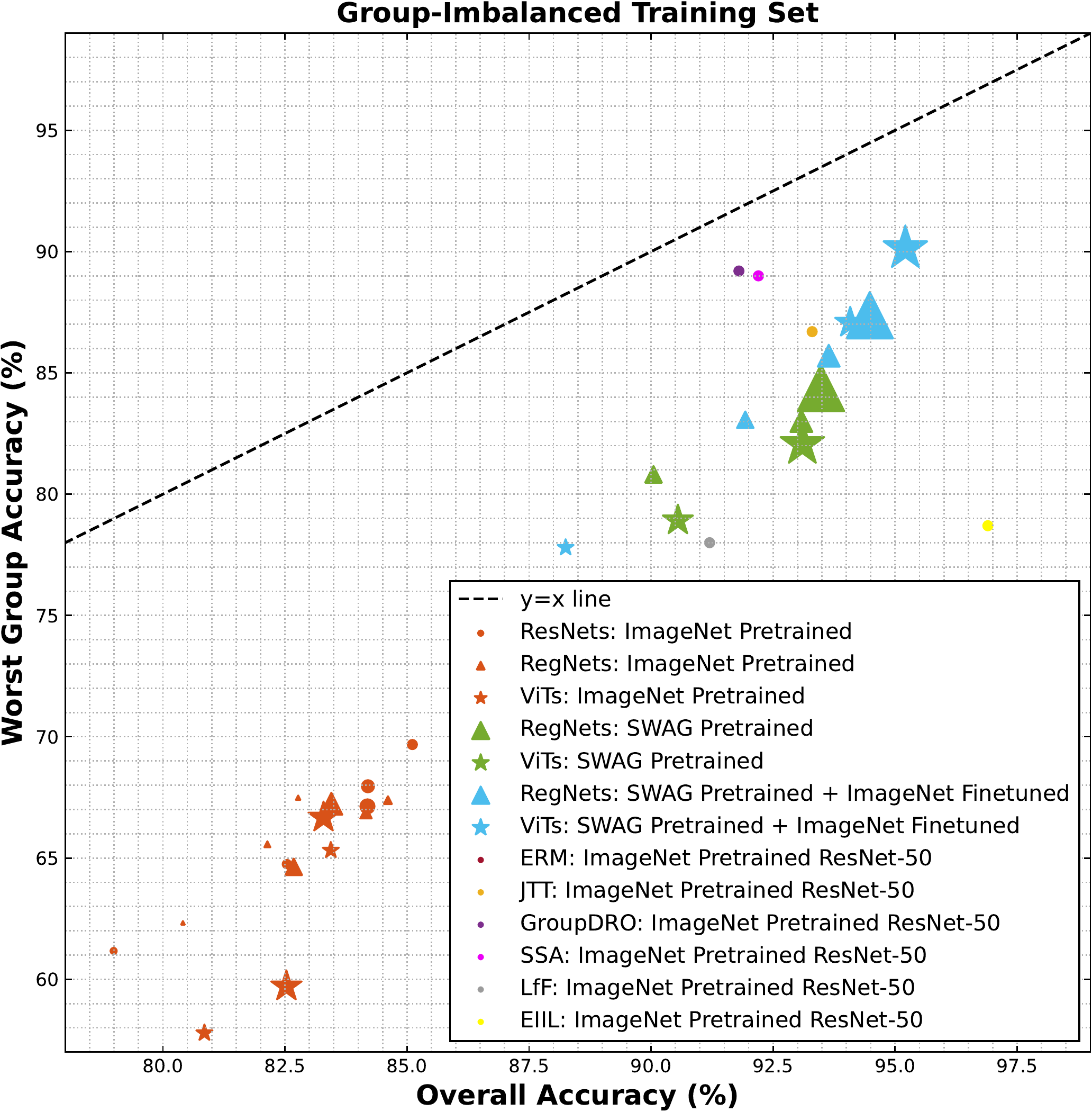}
    \caption{Plot depicting Worst Group Accuracy vs Overall Accuracy (accuracy acorss all images in the dataset) of a linear classifier trained on embeddings extracted from different models (ResNets - $\bullet$, RegNets - $\blacktriangle$, and ViTs - $\bigstar$) pretrained with different datasets (\textcolor{BrickRed}{ImageNet}, \textcolor{ForestGreen}{SWAG}, and \textcolor{Cerulean}{SWAG followed by E2E ImageNet fine-tuning}). Size of a symbol represents the capacity of a model. We observe that embeddings extracted from higher capacity networks pretrained on larger datasets perform better. We also show multiple state-of-the-art spurious debiasing methods like GroupDRO~\cite{sagawa2019distributionally}, JTT~\cite{liu2021just}, SSA~\cite{nam2021spread}, LfF~\cite{nam2020learning}, and EIIL~\cite{creager2021environment}. All these methods train a ResNet-50 ImageNet pretrained model in E2E manner on Waterbirds dataset. We also report ERM results for the same. These results are taken from ~\cite{nam2021spread}.}
    \label{fig:results-a}
\end{figure}

\vspace{-3mm}
\section{Experimental Setup and Dataset}
We experiment with two convolutional neural network architectures, specifically ResNets~\cite{he2016deep} and RegNetYs~\cite{RadosavovicKGHD20}, as well as \textit{Visual Transformers} (ViT)~\cite{DosovitskiyB0WZ21}. We use models with different capacity across these networks which leads to a total of 18 networks. These networks are pre-trained with either ImageNet~\cite{deng2009imagenet} or recently proposed \textit{Supervised Weakly through hashtAGs} (SWAG) dataset~\cite{singh2022revisiting}. For all networks, we use publicly available model weights\footnote{\url{https://pytorch.org/vision/stable/models.html}}. 

We extract embeddings from the above mentioned models without fine-tuning them on the downstream dataset.
We only train a linear classifier on extracted embeddings. For all experiments, we grid search for three different \textit{learning rates} (LR) - 0.01, 0.001, 0.0001, and three different \textit{weight decays} (WD) - 0.0001, 0.00001, 0.000001. We perform all experiments three times with three different random seeds (0, 100, 200) and average their results. We train all models for 20 epochs with a batch size of 32. We measure the performance with overall accuracy and worst-group accuracy. We choose models with the best WGA across the grid searched values of LR and WD.

Waterbirds~\cite{sagawa2019distributionally} is a popular benchmarking dataset used for measuring the effect of debiasing spurious correlations. It combines bird photographs from the Caltech-UCSD Birds-200-2011 (CUB) dataset \cite{WahCUB_200_2011} with image backgrounds from the Places dataset \cite{zhou2017places}. In this dataset, the task is to classify birds as waterbirds or landbirds. The training set contains a total 4,795 image divided into 4 groups: (G1) Waterbirds on water BG - 3,498 images, (G2) Waterbirds on land BG - 184 images, (G3) Landbirds on water BG - 56 images, and (G4) Landbirds on land BG - 1,057 images. G1 and G4 are majority groups, while G2 and G3 are minority groups. The validation and the test set are more balanced across groups. Validation set contains a total 1,199 images (G1 - 467 images, G2 - 467 images, G3 - 133 images, G4 - 133 images), while the test set contains a total 5794 images (G1 - 2,255 images, G2 - 2,255 images, G3 - 642 images, G4 - 642 images).


\section{Experiments and Results}

\noindent\textbf{Effect of embedding extractor model size.}
In Fig. \ref{fig:results-a}, we compare different model architectures pre-trained using same datasets. In particular, we compare 5 different ResNets ($\bullet$), 8 different RegNetYs ($\blacktriangle$), and 4 different ViTs ($\bigstar$). Our experiments show that for each group of networks trained on \textcolor{BrickRed}{ImageNet}, generally higher capacity networks perform marginally better compared to lower capacity networks.  For networks trained on \textcolor{ForestGreen}{SWAG}, we see rather profound effect of model capacity as higher capacity RegNetYs ($\blacktriangle$) and ViTs ($\bigstar$) perform better compared to their \textcolor{BrickRed}{ImageNet} trained counterparts. We observe similar trend for networks trained on \textcolor{Cerulean}{SWAG followed by E2E ImageNet fine-tuning}.\\

\noindent \textbf{Pre-training dataset effect on performance.}
From Fig. \ref{fig:results-a} , we can also compare performance of RegNetY ($\blacktriangle$) and ViT ($\bigstar$) networks which are pre-trained on different datasets. Specifically, we observe that RegNetY networks ($\blacktriangle$) and ViT networks ($\bigstar$) perform better at WGA and OA when pre-trained on \textcolor{Cerulean}{SWAG followed by E2E ImageNet fine-tuning}.
The performance is better compared to networks that are only trained on \textcolor{ForestGreen}{SWAG} or \textcolor{BrickRed}{ImageNet}. We can also observe that network performance becomes closer to the \textit{y=x} line, which shows that the accuracy difference between overall accuracy and worst group accuracy is reduced. \\


\noindent \textbf{Comparison to the state-of-the-art methods.}
The maximum WGA reported in the literature~\cite{sagawa2019distributionally} for the Waterbirds dataset is 89.2\%. 
To achieve this accuracy the authors required group labels and fine-tuned a model end-to-end.
In our experiments, we can achieve higher WGA (90.13\% - \textcolor{Cerulean}{$\bigstar$}) than this by only training a linear classifier on embeddings extracted from a ViT-H-14 network pre-trained on SWAG and followed by E2E finetuning on ImageNet without performing any E2E training on the waterbirds dataset or needing group labels during the training. 

\section{Conclusion}
In this paper, we evaluated the debiasing effect of embeddings extracted from networks with different capacity that are pretrained using different sizes of dataset. We trained a simple linear classifier on top these embeddings. Our experiments reveal that embeddings extracted from a high capacity network, pre-trained using a large dataset has the lower amount of difference between its worst group accuracy and overall accuracy. 

\clearpage
%
%
\bibliographystyle{splncs04}
\bibliography{egbib}



\begin{table}[t]
\centering
\caption{Worst Group Accuracy and Overall Accuracy of a linear classifier trained on embeddings extracted from different ResNet models ($\bullet$) pre-trained on \textcolor{BrickRed}{ImageNet} dataset. We report mean and std across three different runs.}
\begin{tabular}{l|cc}
\Xhline{2\arrayrulewidth}
\multirow{2}{*}{}   & \multicolumn{2}{c}{\cellcolor[HTML]{EFEFEF}\textbf{ImageNet}}                                                                                                                               \\ \cline{2-3} 
                    & \multicolumn{1}{c}{\cellcolor[HTML]{EFEFEF}\textbf{\begin{tabular}[c]{@{}c@{}}Worst Group \\ Accuracy\end{tabular}}} & \cellcolor[HTML]{EFEFEF}\textbf{\begin{tabular}[c]{@{}c@{}}Overall \\ Accuracy\end{tabular}} \\ \Xhline{2\arrayrulewidth}
\cellcolor[HTML]{EFEFEF}\textbf{ResNet-18}  & \multicolumn{1}{c}{61.17 ± 0.65}                                                             & 78.99 ± 0.21                                                         \\ 
\cellcolor[HTML]{EFEFEF}\textbf{ResNet-34}  & \multicolumn{1}{c}{64.75 ± 1.07}                                                             & 82.54 ± 1.57                                                         \\ 
\cellcolor[HTML]{EFEFEF}\textbf{ResNet-50}  & \multicolumn{1}{c}{69.68 ± 0.26}                                                             & 85.11 ± 0.09                                                         \\ 
\cellcolor[HTML]{EFEFEF}\textbf{ResNet-101} & \multicolumn{1}{c}{67.96 ± 0.64}                                                             & 84.20 ± 0.43                                                         \\ 
\cellcolor[HTML]{EFEFEF}\textbf{ResNet-152} & \multicolumn{1}{c}{67.13 ± 0.51}                                                             & 84.19 ± 1.36                                                         \\ \Xhline{2\arrayrulewidth}

\end{tabular}
\end{table}

\begin{table}[t]
\centering
\caption{Worst Group Accuracy and Overall Accuracy of a linear classifier trained on embeddings extracted from different RegNetY models ($\blacktriangle$) pre-trained on \textcolor{BrickRed}{ImageNet} dataset. We report mean and std across three different runs.}
\begin{tabular}{l|cc}
\Xhline{2\arrayrulewidth}
                                                   & \multicolumn{2}{c}{\cellcolor[HTML]{EFEFEF}\textbf{ImageNet}}                                                                                                                                                       \\ \cline{2-3} 
\multirow{-2}{*}{}                                 & \multicolumn{1}{c}{\cellcolor[HTML]{EFEFEF}\textbf{\begin{tabular}[c]{@{}c@{}}Worst Group \\ Accuracy\end{tabular}}} & \cellcolor[HTML]{EFEFEF}\textbf{\begin{tabular}[c]{@{}c@{}}Overall \\ Accuracy\end{tabular}} \\ \Xhline{2\arrayrulewidth}
\cellcolor[HTML]{EFEFEF}\textbf{RegNetY\_400MF}  & \multicolumn{1}{c}{62.23 ± 0.33}                                                                                     & 80.41 ± 0.27                                                                                 \\ 
\cellcolor[HTML]{EFEFEF}\textbf{RegNetY\_800MF}  & \multicolumn{1}{c}{67.49 ± 1.80}                                                                                     & 82.77 ± 0.72                                                                                 \\ 
\cellcolor[HTML]{EFEFEF}\textbf{RegNetY\_1\_6GF} & \multicolumn{1}{c}{65.57 ± 1.70}                                                                                     & 82.14 ± 0.87                                                                                 \\ 
\cellcolor[HTML]{EFEFEF}\textbf{RegNetY\_3\_2GF} & \multicolumn{1}{c}{67.39 ± 0.70}                                                                                     & 84.61 ± 0.71                                                                                 \\ 
\cellcolor[HTML]{EFEFEF}\textbf{RegNetY\_8GF}    & \multicolumn{1}{c}{66.87 ± 4.09}                                                                                     & 84.16 ± 1.05                                                                                 \\ 
\cellcolor[HTML]{EFEFEF}\textbf{RegNetY\_16GF}   & \multicolumn{1}{c}{64.64 ± 1.35}                                                                                     & 82.68 ± 0.52                                                                                 \\ 
\cellcolor[HTML]{EFEFEF}\textbf{RegNetY\_32GF}   & \multicolumn{1}{c}{67.24 ± 0.78}                                                                                     & 83.45 ± 1.19                                                                                 \\ \Xhline{2\arrayrulewidth}
\end{tabular}
\end{table}

\begin{table}[t]
\centering
\caption{Worst Group Accuracy and Overall Accuracy of a linear classifier trained on embeddings extracted from different ViT models ($\bigstar$) pre-trained on \textcolor{BrickRed}{ImageNet} dataset. We report mean and std across three different runs.}
\begin{tabular}{
>{\columncolor[HTML]{EFEFEF}}l |
>{\columncolor[HTML]{FFFFFF}}c 
>{\columncolor[HTML]{FFFFFF}}c }
\Xhline{2\arrayrulewidth}
\cellcolor[HTML]{FFFFFF}                   & \multicolumn{2}{c}{\cellcolor[HTML]{EFEFEF}\textbf{ImageNet}}                                                                                                                                                       \\ \cline{2-3} 
\multirow{-2}{*}{\cellcolor[HTML]{FFFFFF}} & \multicolumn{1}{c}{\cellcolor[HTML]{EFEFEF}\textbf{\begin{tabular}[c]{@{}c@{}}Worst Group \\ Accuracy\end{tabular}}} & \cellcolor[HTML]{EFEFEF}\textbf{\begin{tabular}[c]{@{}c@{}}Overall \\ Accuracy\end{tabular}} \\ \Xhline{2\arrayrulewidth}
\textbf{ViT-B-16}                          & \multicolumn{1}{c}{\cellcolor[HTML]{FFFFFF}65.32 ± 1.21}                                                             & 83.44 ± 0.86                                                                                 \\ 
\textbf{ViT-B-32}                          & \multicolumn{1}{c}{\cellcolor[HTML]{FFFFFF}57.79 ± 1.17}                                                             & 80.85 ± 0.91                                                                                 \\ 
\textbf{ViT-L-16}                          & \multicolumn{1}{c}{\cellcolor[HTML]{FFFFFF}66.67 ± 2.22}                                                             & 83.29 ± 0.74                                                                                 \\ 
\textbf{ViT-L-32}                          & \multicolumn{1}{c}{\cellcolor[HTML]{FFFFFF}59.71 ± 0.96}                                                             & 82.53 ± 0.74                                                                                 \\ \Xhline{2\arrayrulewidth}
\end{tabular}
\end{table}


\begin{table}[t]
\centering
\caption{Worst Group Accuracy and Overall Accuracy of a linear classifier trained on embeddings extracted from different RegNetY models ($\blacktriangle$) pre-trained on \textcolor{ForestGreen}{\textit{SWAG}} dataset. We report mean and std across three different runs.}
\begin{tabular}{
>{\columncolor[HTML]{EFEFEF}}l |
>{\columncolor[HTML]{FFFFFF}}c 
>{\columncolor[HTML]{FFFFFF}}c }
\Xhline{2\arrayrulewidth}
\cellcolor[HTML]{FFFFFF}                   & \multicolumn{2}{c}{\cellcolor[HTML]{EFEFEF}\textbf{SWAG}}                                                                                                                                                       \\ \cline{2-3} 
\multirow{-2}{*}{\cellcolor[HTML]{FFFFFF}} & \multicolumn{1}{c}{\cellcolor[HTML]{EFEFEF}\textbf{\begin{tabular}[c]{@{}c@{}}Worst Group \\ Accuracy\end{tabular}}} & \cellcolor[HTML]{EFEFEF}\textbf{\begin{tabular}[c]{@{}c@{}}Overall \\ Accuracy\end{tabular}} \\ \Xhline{2\arrayrulewidth}
\textbf{RegNetY\_16GF}                   & \multicolumn{1}{c}{\cellcolor[HTML]{FFFFFF}80.82 ± 1.31}                                                             & 90.05 ± 0.70                                                                                 \\ 
\textbf{RegNetY\_32GF}                   & \multicolumn{1}{c}{\cellcolor[HTML]{FFFFFF}83.02 ± 2.02}                                                             & 93.08 ± 0.61                                                                                 \\ 
\textbf{RegNetY\_128GF}                  & \multicolumn{1}{c}{\cellcolor[HTML]{FFFFFF}84.38 ± 0.64}                                                             & 93.48 ± 1.01                                                                                 \\ \Xhline{2\arrayrulewidth}
\end{tabular}
\end{table}

\begin{table}[t]
\centering
\caption{Worst Group Accuracy and Overall Accuracy of a linear classifier trained on embeddings extracted from different ViT models ($\bigstar$) pre-trained on \textcolor{ForestGreen}{\textit{SWAG}} dataset. We report mean and std across three different runs. }
\begin{tabular}{
>{\columncolor[HTML]{EFEFEF}}l |
>{\columncolor[HTML]{FFFFFF}}c 
>{\columncolor[HTML]{FFFFFF}}c }
\Xhline{2\arrayrulewidth}
\cellcolor[HTML]{FFFFFF}                   & \multicolumn{2}{c}{\cellcolor[HTML]{EFEFEF}\textbf{SWAG}}                                                                                                                                                       \\ \cline{2-3} 
\multirow{-2}{*}{\cellcolor[HTML]{FFFFFF}} & \multicolumn{1}{c}{\cellcolor[HTML]{EFEFEF}\textbf{\begin{tabular}[c]{@{}c@{}}Worst Group \\ Accuracy\end{tabular}}} & \cellcolor[HTML]{EFEFEF}\textbf{\begin{tabular}[c]{@{}c@{}}Overall \\ Accuracy\end{tabular}} \\ \Xhline{2\arrayrulewidth}
\textbf{ViT-B-16}                          & \multicolumn{1}{c}{\cellcolor[HTML]{FFFFFF}74.85 ± 0.52}                                                             & 87.10 ± 0.27                                                                                 \\ 
\textbf{ViT-L-16}                          & \multicolumn{1}{c}{\cellcolor[HTML]{FFFFFF}78.92 ± 2.42}                                                             & 90.55 ± 2.38                                                                                 \\ 
\textbf{ViT-H-14}                          & \multicolumn{1}{c}{\cellcolor[HTML]{FFFFFF}82.06 ± 1.40}                                                             & 93.10 ± 0.69                                                                                 \\ \Xhline{2\arrayrulewidth}
\end{tabular}
\end{table}


\begin{table}[t]
\centering
\caption{Worst Group Accuracy and Overall Accuracy of a linear classifier trained on embeddings extracted from different RegNetY models ($\blacktriangle$) pre-trained on \textcolor{Cerulean}{\textit{SWAG} followed by \textit{E2E} ImageNet fine-tuning}. We report mean and std across three different runs.}
\begin{tabular}{
>{\columncolor[HTML]{EFEFEF}}l |
>{\columncolor[HTML]{FFFFFF}}c 
>{\columncolor[HTML]{FFFFFF}}c }
\Xhline{2\arrayrulewidth}
\cellcolor[HTML]{FFFFFF}                   & \multicolumn{2}{c}{\cellcolor[HTML]{EFEFEF}\textbf{SWAG+ImageNet-FT}}                                                                                                                                               \\ \cline{2-3} 
\multirow{-2}{*}{\cellcolor[HTML]{FFFFFF}} & \multicolumn{1}{c}{\cellcolor[HTML]{EFEFEF}\textbf{\begin{tabular}[c]{@{}c@{}}Worst Group \\ Accuracy\end{tabular}}} & \cellcolor[HTML]{EFEFEF}\textbf{\begin{tabular}[c]{@{}c@{}}Overall \\ Accuracy\end{tabular}} \\ \Xhline{2\arrayrulewidth}
\textbf{RegNetY\_16GF}                   & \multicolumn{1}{c}{\cellcolor[HTML]{FFFFFF}83.07 ± 1.58}                                                             & 91.93 ± 0.38                                                                                 \\ 
\textbf{RegNetY\_32GF}                   & \multicolumn{1}{c}{\cellcolor[HTML]{FFFFFF}85.71 ± 1.40}                                                             & 93.64 ± 0.79                                                                                 \\ 
\textbf{RegNetY\_128GF}                  & \multicolumn{1}{c}{\cellcolor[HTML]{FFFFFF}87.38 ± 1.96}                                                             & 94.48 ± 0.59                                                                                 \\ \Xhline{2\arrayrulewidth}
\end{tabular}
\end{table}

\begin{table}[t]
\centering
\caption{Worst Group Accuracy and Overall Accuracy of a linear classifier trained on embeddings extracted from different ViT models ($\bigstar$) pre-trained on \textcolor{Cerulean}{\textit{SWAG} followed by \textit{E2E} ImageNet fine-tuning}. We report mean and std across three different runs.}
\begin{tabular}{
>{\columncolor[HTML]{EFEFEF}}l |
>{\columncolor[HTML]{FFFFFF}}c 
>{\columncolor[HTML]{FFFFFF}}c }
\Xhline{2\arrayrulewidth}
\cellcolor[HTML]{FFFFFF}                   & \multicolumn{2}{c}{\cellcolor[HTML]{EFEFEF}\textbf{SWAG+ImageNet-FT}}                                                                                                                                               \\ \cline{2-3} 
\multirow{-2}{*}{\cellcolor[HTML]{FFFFFF}} & \multicolumn{1}{c}{\cellcolor[HTML]{EFEFEF}\textbf{\begin{tabular}[c]{@{}c@{}}Worst Group \\ Accuracy\end{tabular}}} & \cellcolor[HTML]{EFEFEF}\textbf{\begin{tabular}[c]{@{}c@{}}Overall \\ Accuracy\end{tabular}} \\ \Xhline{2\arrayrulewidth}
\textbf{ViT-B-16}                          & \multicolumn{1}{c}{\cellcolor[HTML]{FFFFFF}77.80 ± 1.57}                                                             & 88.25 ± 1.02                                                                                 \\ 
\textbf{ViT-L-16}                          & \multicolumn{1}{c}{\cellcolor[HTML]{FFFFFF}87.07 ± 1.14}                                                             & 94.08 ± 0.75                                                                                 \\ 
\textbf{ViT-H-14}                          & \multicolumn{1}{c}{\cellcolor[HTML]{FFFFFF}90.13 ± 0.91}                                                             & 95.21 ± 0.45                                                                                 \\ \Xhline{2\arrayrulewidth}
\end{tabular}
\end{table}


\begin{table}[t]
\centering
\caption{Worst Group Accuracy and Overall Accuracy for different state-of-the-art debiasing spurious correlation methods. We report results for ERM, LfF \cite{nam2020learning}, EIIL \cite{creager2021environment}, JTT \cite{liu2021just}, SSA \cite{nam2021spread}, and GroupDRO \cite{sagawa2019distributionally} as reported in \cite{nam2020learning}. We are able to achieve better results compared to any of the about mentioned methods by training a linear classifier on the top of embeddings extacted from a ViT-H-14 network pre-trained on SWAG followed by E2E ImageNet finetuning. }
\label{tab:my-table}
\resizebox{0.95\linewidth}{!}{
\begin{tabular}{
>{\columncolor[HTML]{EFEFEF}}l |cccc|
>{\columncolor[HTML]{FFFFFF}}c 
>{\columncolor[HTML]{FFFFFF}}c }
\Xhline{2\arrayrulewidth}
\cellcolor[HTML]{EFEFEF}              & \cellcolor[HTML]{EFEFEF}\textbf{Network} & \cellcolor[HTML]{EFEFEF}\textbf{\begin{tabular}[c]{@{}c@{}}Pre-training\\ Data\end{tabular}} & \cellcolor[HTML]{EFEFEF}\textbf{\begin{tabular}[c]{@{}c@{}}Waterbirds\\ Fine-tuning\end{tabular}} & \cellcolor[HTML]{EFEFEF}\textbf{\begin{tabular}[c]{@{}c@{}}Amount of \\ Group label used\end{tabular}} & \cellcolor[HTML]{EFEFEF}\textbf{\begin{tabular}[c]{@{}c@{}}Worst Group \\ Accuracy\end{tabular}} & \cellcolor[HTML]{EFEFEF}\textbf{\begin{tabular}[c]{@{}c@{}}Overall \\ Accuracy\end{tabular}} \\ \Xhline{2\arrayrulewidth}
\cellcolor[HTML]{EFEFEF}\textbf{ERM}  & ResNet-50                                & ImageNet                                                                                     & \Checkmark                                                                                        & validation set                                                                                         & \cellcolor[HTML]{FFFFFF}72.6                                                                     & \cellcolor[HTML]{FFFFFF}97.3                                                                 \\ 
\cellcolor[HTML]{EFEFEF}\textbf{LfF}  & ResNet-50                                & ImageNet                                                                                     & \Checkmark                                                                                        & validation set                                                                                         & \cellcolor[HTML]{FFFFFF}78.0                                                                     & \cellcolor[HTML]{FFFFFF}91.2                                                                 \\ \
\cellcolor[HTML]{EFEFEF}\textbf{EIIL} & ResNet-50                                & ImageNet                                                                                     & \Checkmark                                                                                        & validation set                                                                                         & \cellcolor[HTML]{FFFFFF}78.7                                                                     & \cellcolor[HTML]{FFFFFF}96.9                                                                 \\ 
\textbf{JTT}                          & ResNet-50                                & ImageNet                                                                                     & \Checkmark                                                                                        & validation set                                                                                         & 86.7                                                                                             & 93.3                                                                                         \\ 
\textbf{SSA}                          & ResNet-50                                & ImageNet                                                                                     & \Checkmark                                                                                        & validation set                                                                                         & 89.0 ± 0.55                                                                                      & 92.2 ± 0.87                                                                                  \\ 
\textbf{GroupDRO}                     & ResNet-50                                & ImageNet                                                                                     & \Checkmark                                                                                        & \begin{tabular}[c]{@{}c@{}}training and \\ validation set\end{tabular}                                                                            & 89.2 ± 0.18                                                                                      & 91.8 ± 0.48                                                                                  \\ \Xhline{2\arrayrulewidth}
\textbf{Ours-ResNet-50}               & ResNet-50                                & ImageNet                                                                                     & \XSolidBrush                                                                                      & validation set                                                                                         & 69.7 ± 0.26                                                                                      & 85.1 ± 0.09                                                                                  \\ 
\textbf{Ours-ViT-H-14}                & ViT-H-14                                 & SWAG+ImageNet                                                                                & \XSolidBrush                                                                                      & validation set                                                                                         & 90.1 ± 0.91                                                                                      & 95.2 ± 0.45                                                                                  \\ \Xhline{2\arrayrulewidth}
\end{tabular}
}
\end{table}

\end{document}